\documentclass[letterpaper]{article} 
\usepackage{aaai25}  
\usepackage{times}  
\usepackage{helvet}  
\usepackage{courier}  
\usepackage[hyphens]{url}  
\usepackage{graphicx} 
\usepackage{natbib}  
\usepackage{caption} 
\usepackage{algorithm}
\usepackage{algorithmic}
\usepackage{newfloat}
\usepackage{listings}

\usepackage{amsmath, amsfonts, amssymb}
\usepackage{array}
\usepackage{booktabs}
\usepackage{enumitem}
\usepackage{graphicx}
\usepackage{microtype}
\usepackage{multirow}
\usepackage{nicefrac}
\usepackage{printlen}
\usepackage{soul}
\usepackage{svg}
\usepackage{lipsum}
\usepackage{xcolor}
\usepackage{lipsum}

\usepackage{framed}
\colorlet{shadecolor}{blue!20}

\newcommand{\com}[1]{\iffalse #1 \fi}%

\newcommand{\noimage}{%
  \setlength{\fboxsep}{-\fboxrule}%
  \fbox{\phantom{\rule{100pt}{100pt}}File missing\phantom{\rule{100pt}{100pt}}}
}
\let\includegraphicsoriginal\includegraphics
\renewcommand{\includegraphics}[2][width=\textwidth]{\IfFileExists{#2}{\includegraphicsoriginal[#1]{#2}}{\noimage}}

\newcolumntype{+}{>{\global\let\currentrowstyle\relax}}
\newcolumntype{^}{>{\currentrowstyle}}

\newenvironment{aequation}
{\begin{equation} \begin{aligned}}
{\end{aligned} \end{equation}}

\emergencystretch=\maxdimen
\everypar{\looseness=-1 }

\newcounter{descriptcount}

\definecolor{Overlapped}{RGB}{180, 22 , 0   }
\definecolor{Fractured}{RGB}{23 , 77 , 127 }
\definecolor{Clustered}{RGB}{55 , 144, 48  }
\definecolor{hl}{RGB}{255,20,147}

\definecolor{yellow}{HTML}{FF7F00} 
\definecolor{green}{HTML}{4D06AF}
\definecolor{pink}{HTML}{93256B}

\usepackage{tikz}

\newrobustcmd*{\mysquare}[1]{\tikz{\filldraw[draw=#1,fill=#1] (0,0)
rectangle (0.2cm,0.2cm);}}

\newrobustcmd*{\mycircle}[1]{\tikz{\filldraw[draw=#1,fill=#1] (0,0) circle [radius=0.1cm];}}

\newrobustcmd*{\mytriangle}[1]{\tikz{\filldraw[draw=#1,fill=#1] (0,0) --
(0.2cm,0) -- (0.1cm,0.2cm);}}

\newcommand{\Fractured}{{$\bigstar$}}
\newcommand{\Overlapped}{{$\clubsuit$}}
\newcommand{\Clustered}{{$\spadesuit$}}

\newcommand{\PatternA}{{\textbf{Pattern~A}~(\Fractured)}}
\newcommand{\PatternB}{{\textbf{Pattern~B}~(\Overlapped)}}
\newcommand{\PatternC}{{\textbf{Pattern~C}~(\Clustered)}}

\newcommand{\overlapped}{{\textbf{Overlapped}}}
\newcommand{\clustered}{{\textbf{Clustered}}}
\newcommand{\fractured}{{\textbf{Fractured}}}

\sethlcolor{hl}
\makeatletter
\def\SOUL@hlpreamble{%
    \setul{\dp\strutbox}{\dimexpr\ht\strutbox+\dp\strutbox\relax}%
    \let\SOUL@stcolor\SOUL@hlcolor
    \SOUL@stpreamble
}
\makeatother

{\color{hl}}%
{}

\definecolor{Gray}{gray}{0.85}
\definecolor{cite}{HTML}{53769A}
\definecolor{ref}{HTML}{379030}
\definecolor{lightcornflowerblue}{rgb}{0.6, 0.81, 0.93}
\definecolor{lightkhaki}{rgb}{0.94, 0.9, 0.55}
\definecolor{lightmauve}{rgb}{0.86, 0.82, 1.0}
\definecolor{lightgreen}{rgb}{0.56, 0.93, 0.56}

\definecolor{royalpurple}{RGB}{207,199,216}
\definecolor{forestgreen}{RGB}{202,225,204}

\usepackage{collcell}
\usepackage{xstring}
\usepackage{datatool}
\usepackage{booktabs}
\usepackage{environ}

\newcounter{CurrentRow}
\newcounter{CurrentColumn}
\newtoggle{DoneWithFirstRow}

\newcommand*{\FirstColumn}[1]{%
    \IfEq{\arabic{CurrentColumn}}{0}{%
        \global\togglefalse{DoneWithFirstRow}%
        \setcounter{CurrentRow}{1}
    }{%
        \global\toggletrue{DoneWithFirstRow}%
        \stepcounter{CurrentRow}%
    }%
    \setcounter{CurrentColumn}{0}%
    \NewData{#1}%
}
\newcommand*{\NewData}[1]{%
    \dtlexpandnewvalue%
    \stepcounter{CurrentColumn}%
    \iftoggle{DoneWithFirstRow}{%
        \dtlgetrow{TransposedTabularDB}{\arabic{CurrentColumn}}%
        \dtlappendentrytocurrentrow{\Alph{CurrentRow}}{#1}%
        \dtlrecombine%
    }{%
        \DTLnewrow{TransposedTabularDB}%
        \DTLnewdbentry{TransposedTabularDB}{\Alph{CurrentRow}}{#1}%
    }%
}%

\newcolumntype{F}{>{\collectcell\FirstColumn}c<{\endcollectcell}}
\newcolumntype{C}{>{\collectcell\NewData}{c}<{\endcollectcell}}

\newtoggle{EncounteredDataRow}

\newsavebox{\TempBox}
\DTLnewdb{TransposedTabularDB}

\NewEnviron{Ttabular}[1]{%
    \setcounter{CurrentColumn}{0}%
    \global\togglefalse{EncounteredDataRow}%
    \savebox{\TempBox}{%
        \begin{tabular}{FCCCCCC}
            \BODY%
        \end{tabular}%
    }%
    \begin{tabular}{#1}\toprule%
    \DTLforeach*{TransposedTabularDB}{\Aa=A, \Ba=B, \Ca=C}{%
      \DTLiffirstrow{}{\\\midrule}%
        \Aa & \Ba & \Ca %
    }\\\bottomrule%
    \end{tabular}%
}%

\newcolumntype{H}{>{\setbox0=\hbox\bgroup}c<{\egroup}@{}}

\usepackage{mathabx,graphicx}

\def\Figref#1{Figure~\ref{#1}}

\def\Tableref#1{Table~\ref{#1}}

\def\eqref#1{equation~\ref{#1}}
\def\Eqref#1{Equation~\ref{#1}}

\def\1{\bm{1}}

\DeclareMathAlphabet{\mathsfit}{\encodingdefault}{\sfdefault}{m}{sl}
\SetMathAlphabet{\mathsfit}{bold}{\encodingdefault}{\sfdefault}{bx}{n}

\DeclareMathOperator*{\argmin}{arg\,min}

\DeclareCaptionStyle{ruled}{labelfont=normalfont,labelsep=colon,strut=off} 
\floatstyle{ruled}
\newfloat{listing}{tb}{lst}{}
\floatname{listing}{Listing}

\usepackage{color}

\title{Linking Robustness and Generalization:\\
A k* Distribution Analysis of Concept Clustering in Latent Space for Vision Models
}
\author {
    Shashank Kotyan\textsuperscript{\rm 1},
    Pin-Yu Chen\textsuperscript{\rm 2},
    Danilo VAsconcellos Vargas\textsuperscript{\rm 1}
}
\affiliations {
    \textsuperscript{\rm 1}Laboratory of Intelligent Systems, Kyushu University\\
    \textsuperscript{\rm 2}IBM Research\\
}

\begin{document}

\maketitle

\begin{abstract}

Most evaluations of vision models use indirect methods to assess latent space quality. 
These methods often involve adding extra layers to project the latent space into a new one. 
This projection makes it difficult to analyze and compare the original latent space.
This article uses the k* Distribution, a local neighborhood analysis method, to examine the learned latent space at the level of individual concepts, which can be extended to examine the entire latent space. 
We introduce skewness-based true and approximate metrics for interpreting individual concepts to assess the overall quality of vision models' latent space. 
Our findings indicate that current vision models frequently fracture the distributions of individual concepts within the latent space. 
Nevertheless, as these models improve in generalization across multiple datasets, the degree of fracturing diminishes. 
A similar trend is observed in robust vision models, where increased robustness correlates with reduced fracturing. 
Ultimately, this approach enables a direct interpretation and comparison of the latent spaces of different vision models and reveals a relationship between a model’s generalizability and robustness. 
Results show that as a model becomes more general and robust, it tends to learn features that result in better clustering of concepts.


\end{abstract}

\section{Introduction}

The rapid advancements in computer vision and deep learning have led to the development of powerful vision models capable of extracting intricate features from visual data \cite{radford2021learning,jia2021scaling,cherti2023reproducible}. 
These models are central to various applications, from object recognition to image generation.
Typically, their generalizability is measured through zero-shot classification performance \cite{wortsman2022robust}, making the evaluation of vision models indirect. 
However, these evaluations often rely on a projection of the learned latent space, which may not fully capture the quality or nuances of the underlying representations and offer little insight into improving them.

Understanding the structure and quality of a latent space is crucial for gaining insights into how vision models process and organize visual information. 
Traditional methods like t-SNE \cite{maaten2008visualizing} and UMAP \cite{2018arXivUMAP} offer visualizations of this high-dimensional space by reducing its dimensions. 
While these methods provide insights about a latent space, they are less effective when comparing multiple latent spaces. 
As vision models become more sophisticated, there is a growing need for methods that offer a more detailed and interpretable analysis of the latent space beyond mere visual inspection.

In this context, we focus on local neighborhood structures within the high-dimensional latent space analyzed using the k* distribution proposed by \cite{kotyan2023k} for evaluating and comparing the latent spaces. 
This method preserves the neighborhood information, similar to t-SNE \cite{maaten2008visualizing} and UMAP \cite{2018arXivUMAP}, in the local neighborhood of samples and directly examines the distribution and clustering of individual concepts. 
By doing so, k* Distributions offers a more nuanced understanding of the relationships within the latent space and provides objective comparisons between the latent spaces.

The k* distribution assesses the index of the nearest neighbor from a different concept (class), offering insights into the cohesion and fracture in distributions belonging to similar concepts.
This enables meaningful comparisons between the distribution of samples belonging to different classes and multiple latent spaces. 
Moreover, it is complementary to the existing analyses available.  
By providing the understanding about the structure of distribution of samples in the latent space, this approach augments the insights from dimensionality reduction visualizations or projection-based evaluation, making the interpretability and comparison of the latent spaces easier. 

\begin{figure*}[!t]
\centering
\includegraphics[width=0.99\textwidth]{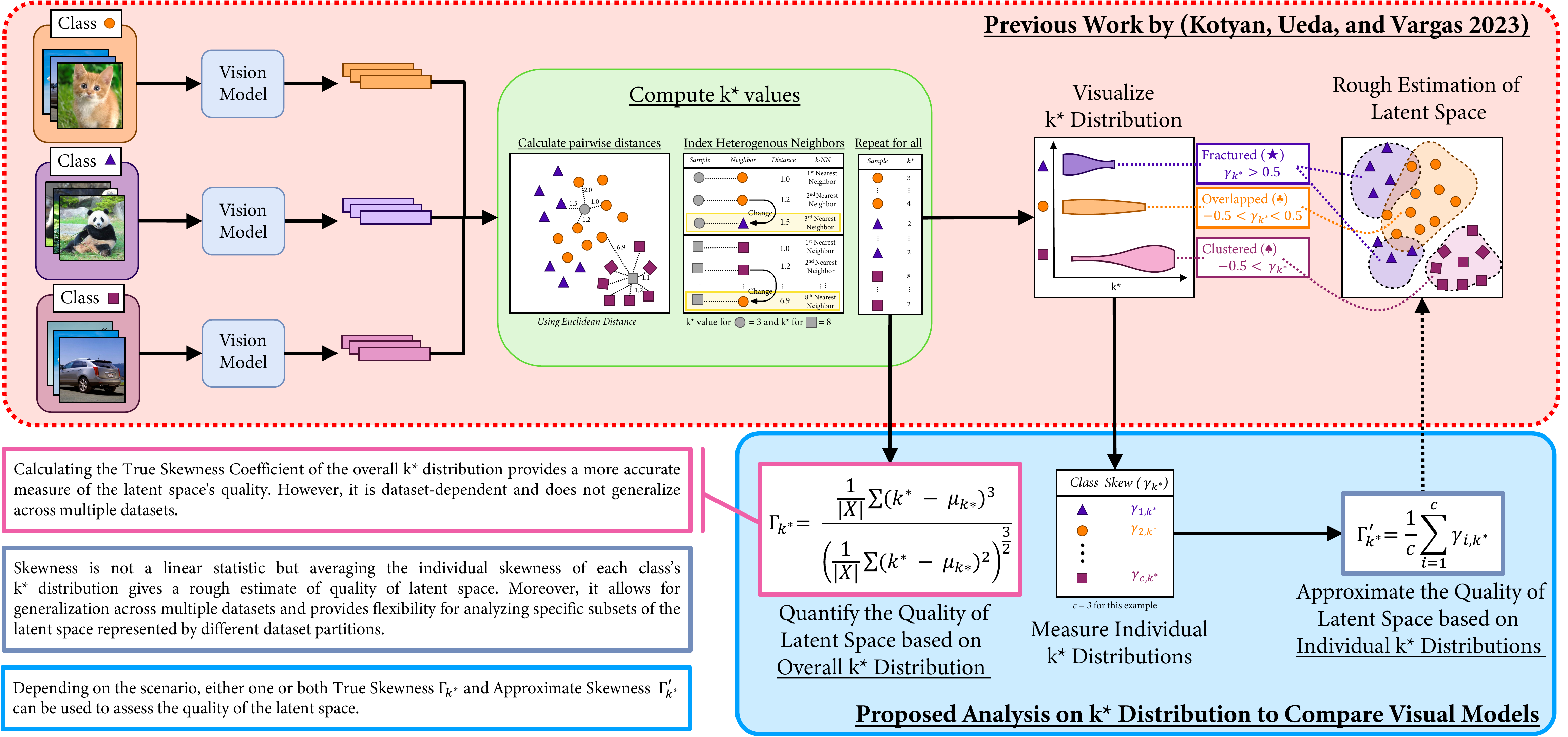}
\caption{
    Overview of the framework to create and analyze {k*~Distribution}.
    We use the learned features of a vision model to compute k* values of individual evaluated samples and then compute the {k*~distribution} for a particular concept (class). 
    We then evaluate the entire learned latent space by quantifying the quality of latent spaced using either True Skewness Coefficient $\Gamma_{k^*}$ based on Overall k* distribution or an Approximate Skewness Coefficient $\Gamma^{'}_{k^*}$ based on averaging the skew of individual k* distributions.
}
\label{fig:framework}
\end{figure*}

\subsection{Contributions}

\begin{description}[style=nextline, leftmargin=*]

\item[Quantification of Quality of Latent Space:]
We quantify the quality of latent space based on the Skewness coefficient of k* Distribution as described in \Figref{fig:framework}.
We derive a True Skewness Coefficient $\Gamma_{k^*}$ as a more accurate measure of latent space's quality, and we derive an Approximate Skewness Coefficient $\Gamma^{'}_{k^*}$ that can be used across concepts from multiple datasets.

\item[Large Scale Study of Robust Vision Models:] 
We compare the different types of robust models available at RobustBench Library \cite{croce2020robustbench} and note that more robust models have less degree of fracturing in the latent space. 
This indicates that as the models become more robust, they can cluster the individual concepts correctly, narrowing the tradeoff between accuracy and robustness. 

\item[Large Scale Study of CLIP-based Vision Models:] 
We compare the different pre-trained CLIP models available at OpenCLIP Library \cite{ilharco_gabriel_2021_5143773} and observe that there is a visible less degree of fracturing in the latent space by CLIP models that are better at generalizability for most evaluated datasets.
This suggests that as models become better at generalizing to other datasets, they also become better at clustering the individual concepts. 

\end{description}

\section{Related Works}

\subsection{Analyzing the Latent Space of Neural Networks}

Analyzing the latent space of deep neural networks poses a significant challenge due to the high-dimensional nature of the features. 
To address this, researchers have developed various dimensionality reduction techniques that allow for the visualization and interpretation of these complex spaces in lower dimensions, typically 2D or 3D. 
Among the most widely used techniques are t-SNE \cite{maaten2008visualizing} and UMAP \cite{2018arXivUMAP}, which are used for their ability to preserve local and global structures, respectively. 
These techniques are part of a broader family of methods designed to make the high-dimensional latent space more interpretable by projecting it into a more manageable form, including classical approaches like PCA \cite{hotelling1933analysis} and MDS \cite{kruskal1964multidimensional}, as well as more recent methods like Diffusion Maps \cite{coifman2006diffusion} and TriMap \cite{amid2022TriMap}.

Interpretation using dimensionality reduction techniques largely depends on the organization of the latent space. 
When the latent space is well-structured, and the encoded information aligns with meaningful patterns, these visualizations can be highly effective, providing insights that correlate with established interpretations. 
However, where the latent space lacks a clear structure, the utility of these methods diminishes. 
Without predefined organization, the resulting plots often appear as amorphous clusters, providing little actionable insight \cite{sivaraman2022Emblaze}.

Beyond dimensionality reduction, other approaches focus on visualizing the interactions between neural network features and the latent space. 
By analyzing the activation patterns of hidden units in response to specific inputs, researchers can gain insights into which features are emphasized by the network \cite{visualizing_activation_1, visualizing_activation_2}. 
Tools like Activation Atlas \cite{activation-atlas} offers a way to explore how combinations of features are represented, further illuminating the structure of the learned latent space.

However, these visualization techniques often involve complex hyperparameter tuning, making it difficult to achieve consistent and fair comparisons across different latent spaces, especially when comparing vision models with varying dimensionalities. 
The interpretability of these visualizations becomes increasingly complicated when multiple models are involved, as the differences in their latent space structures can lead to incomparable results. 
This complexity underscores the need for more robust and interpretable methods to analyze and compare latent spaces across models \cite{gleicher2011Visual,arendt2020Parallel,cutura2020Comparing,boggust2022Embedding,sivaraman2022Emblaze}.

\subsection{Evaluating the Latent Space of Neural Networks}

Recent research has focused on evaluating the effectiveness of vision models through their performance in downstream tasks, particularly zero-shot transfer classification across multiple datasets like proposed in VTAB (collection of 19 different image-classification datasets) \cite{zhai2019large} and ELEVATER  (collection of 20 different image-classification datasets)\cite{li2022elevater}. 
In these studies, a vision model's performance on a variety of datasets is used as a proxy for measuring its generalization capabilities, with improved results across diverse datasets interpreted as evidence of broader visual concept coverage.
This performance is often evaluated using Zero-shot classification which is typically categorized into two main types:
a) (Traditional) Class-Level Zero-Shot, which measures how well a vision model generalizes to unseen object categories, and 
b) (More recent) Task-Level Zero-Shot, which assesses the vision model's ability to generalize to entirely new datasets. 

While these evaluations provide useful insights, they are often evaluated indirectly using projection of the latent space and do not directly measure the intrinsic quality of the vision model itself.
Language-Free Vision Models, for instance, rely solely on a vision model that produces feature vectors from images, followed by a randomly initialized linear layer acting as the classifier \cite{dosovitskiy2020image}. 
Although these can be adapted to different tasks, this often depends on external factors, such as the effectiveness of the added classifier, rather than the model's inherent capability to capture visual information.
On the other hand, Language-Augmented Vision Models integrate image and text encoders, projecting features into a shared space \cite{radford2021learning,jia2021scaling}. 
This approach enables zero-shot learning by comparing image features with averaged text features representing different categories. 
However, the success of such models may be more attributable to the alignment between visual and textual representations than to the vision models' standalone performance.

Moreover, adaptation techniques like random-initialized adaptation, where a linear layer is added to a pre-trained vision model without using language features \cite{kornblith2019better}, and language-initialized adaptation, which can either initialize the linear layer with text features or directly integrate visual and text features into a single projection, further complicate the evaluation process \cite{zhou2022learning,wortsman2022robust}. 
These methods introduce additional variables that can obscure the true quality of the image encoder, making it difficult to isolate and assess its effectiveness independently of the adaptation process.

In summary, existing evaluation methods provide valuable insights into the performance of vision models across various tasks. However, these methods often rely on indirect measures and are influenced by factors unrelated to the models' inherent capabilities. Conversely, dimensionality reduction techniques offer a more direct interpretation of the learned latent space but pose challenges when comparing multiple latent spaces. In this article, we evaluate and analyze the latent spaces of vision models using the k* distribution, aiming to enhance the interpretability of dimensionality reduction techniques and facilitate meaningful comparisons across different tasks.

\section{k* Distribution for Latent Space Analysis}

The k* distribution proposed by \cite{kotyan2023k} provides a robust method for analyzing the structure of hyperdimensional latent spaces learned by Vision Models, focusing on local neighborhood dynamics. 
This approach is instrumental in understanding how samples and clusters are distributed within the latent space by associating them with their respective concepts (classes). 
By analyzing k* distribution, one can gain valuable insights into the patterns and formations of clusters and their underlying structure within the latent space.

At the core of this methodology is the k* value, which represents the index of the  k\textsuperscript{th} nearest neighbor that belongs to a different concept (class) than the test sample. 
This value measures the neighborhood uniformity; 
a high k* value suggests that a sample is surrounded by many neighbors from the same class, indicating a well-formed homogeneous cluster. 
On the other hand, a low k* value points to the proximity of a neighbor from a different class, signaling potential overlap or a fragmented neighborhood.
This approach assesses the homogeneity and cohesion of class clusters and helps identify whether they are concentrated or dispersed across multiple regions within the latent space.

\subsection{Mathematical Framework}
Consider a set of sample-label pairs $X$:
$(x_1, Y_1), (x_2, Y_2), ..., (x_n, Y_n)$
where $x$ represents the input samples and $Y$ denotes their corresponding labels. 
For a given concept (class) $c$, let $S_c$ denote the set of all samples with the same label $c$:
\begin{aequation}
    S_c = \{x_i \mid \forall x_i \in X ~\text{such that}~ Y_{i} = c \}
    \label{eq:S}
\end{aequation}

The $k^{\text{th}}$ nearest neighbour $x_{k}^{p}$ of a sample $x_p$ is defined as,
\begin{aequation}
x_{k}^{p} = x_q \quad \text{where,} \quad q \in \argmin_{x_q \in P_i}~ \text{distance}(x_q, x_p) \\
\text{such that} \quad P_i = X - \{ x_{j}^{p} \mid \forall j < i\}
\end{aequation}
where $\text{distance}(a,b)$ is the distance between two samples $a$ and $b$.
Using this, we can construct a sorted local neighborhood space $N_p$ of sample $(x_p)$ as defined:
\begin{aequation}
N_p = (x_{0}^{p}, x_{1}^{p} \ldots x_{n}^{p})
\end{aequation}
here, $\text{distance}(x_{i}^{p}, x_p) < \text{distance}(x_{j}^{p}, x_p)$, where $i < j$.

The k* value of a test sample $(x_p, Y_p)$ is the index of the nearest neighbor that has a different label than $Y_p$, formally defined as:
\begin{aequation}
\text{k}^{*}_p = \argmin_{(x_p, Y_p)} \{ x_{i}^{p} \mid x_{i}^{p} \in N_p, ~ Y_{i}^{p} \ne Y_p \},
\end{aequation}
where $i$ is the index of the nearest neighbor,
$Y_p$ is the label of test sample $x_p$ and
$Y_{i}^{p}$ is the label of the nearest neighbor (sample) $x_{i}^{p}$ that differs compared to label $Y_p$.
Thus, the {k*~distribution} $\text{k}^{*}(\cdot)$ of concept (class) $c$ can be written as,
\begin{aequation}
\text{k}^{*}(S_c) = \left\{ \frac{\text{k}^{*}_p}{\vert S_c \vert} \mid \forall x_p \in S_c \right\},
\end{aequation}
here $\vert S_c \vert$ is the number of samples of concept (class) $c$.
Finally, the Skewness Coefficient of {k*~distribution} ($\gamma_{k^*}$) can be defined as
\begin{aequation}
\gamma_{k^*} = \frac{ \frac{1}{\vert S \vert} \sum (\text{k}^{*}(S) - \mu_{k^*})^3}{ \left(\frac{1}{\vert S \vert} \sum (\text{k}^{*}(S) - \mu_{k^*})^2 \right)^{3/2}}
\label{eq:gamma}
\end{aequation}%
where $\mu_{k^*}$ is the mean of k* distribution. 
It measures the asymmetry of the {k*~distribution} about its mean $\mu_{k^*}$.
Positive skewness indicates a distribution skewed towards lower k* values, indicating fracturing, while negative skewness indicates a distribution skewed towards higher k* values, indicating clustering.

We can denote, the skew of k* distribution $\gamma_{k^*}$ for each $i^{\text{th}}$ concept (class) available in our sample-label pairs $X$ as $\gamma_{i, k^*}$. 
In this case, the $S$ in the equation corresponds to the subset of dataset points as defined in \Eqref{eq:S}. 
Using the individual k* distributions $\gamma_{i, k^*}$, we can define the Approximate Skewness Coefficient $\Gamma^{'}_{k^*}$  as, 
\begin{aequation}
\Gamma^{'}_{k^*} = \frac{1}{c} \sum_{i=1}^{c} \gamma_{i, k^*}
\end{aequation}

For measuring the True Skewness Coefficient $\Gamma_{k^*}$ using the overall k* distribution computed for the entire evaluated sample-label pairs $X$, \Eqref{eq:gamma} can be rewritten as,  
\begin{aequation}
\Gamma_{k^*} = \frac{ \frac{1}{\vert X \vert} \sum (\text{k}^{*}(X) - \mu_{k^*})^3}{ \left(\frac{1}{\vert X \vert} \sum (\text{k}^{*}(X) - \mu_{k^*})^2 \right)^{3/2}}
\end{aequation}

\textbf{Note:} The Approximate Skewness Coefficient $\Gamma^{'}_{k^*}$ is not a true metric as the skew coefficient $\gamma_{k^*}$ is not a linear statistic. 
However, it is an approximation assuming that the individual $\gamma_{i, k^*}$ are independent and have low variance. 
Further, Central Limit Theorem (CLT) and the Law of Large Numbers ensure that this Approximate Skewness Coefficient $\Gamma^{'}_{k^*}$ will converge to the True Skewness Coefficient $\Gamma_{k^*}$ as the number of concepts (classes) increase, provided that the individual $\gamma_{i, k^*}$ remain independent. 
This assumption that the individual $\gamma_{i, k^*}$ are independent is also useful when we want to compare the concepts from different datasets, i.e., sample-label pairs $X$.

\subsection{Patterns in Latent Space Distribution}

\begin{description}[style=nextline, leftmargin=*]

    \item[\PatternA~\fractured~distribution of samples:]
    In this latent space configuration, multiple clusters of testing samples are observable; each separated in the latent space.
    Consequently, most points exhibit low k* values, as they belong to smaller clusters.
    Conversely, no points display high k* values, given the presence of points from another class distribution situated between the various sub-clusters of the testing class.
    The {k*~distribution} for this clustered distribution of samples in latent space is markedly positively skewed ($\gamma_{k^*} > 0.5$), i.e., skewed towards lower k* values.

    \item[\PatternB~\overlapped~distribution of samples:]
    This latent space configuration represents the scenario when samples from two or more classes overlap.
    Consequently, some points possess low k* values, suggesting their location in the overlapping region, while others have high k* values, signifying their deep embedding within a class cluster.
    Due to the diverse distribution of samples in this latent space, the {k*~distribution} appears nearly uniform ($-0.5 < \gamma_{k^*} < 0.5$).

    \item[\PatternC~\clustered~distribution of samples:]
    A homogeneous cluster of testing samples is prevalent in this latent space arrangement.
    As a result, most points boast high k* values, indicating their deep placement within the cluster.
    Simultaneously, some points may exhibit low k* values as they reside on the cluster's periphery; these edge samples might be closer to points from another class distribution than the majority within the cluster.
    Owing to this concentrated distribution of samples, the {k*~distribution} for this clustered distribution of samples in latent space is strongly negatively biased ($\gamma_{k^*} < -0.5$), i.e., skewed towards higher k* values, symbolizing a dense cluster.

\end{description}

\begin{figure*}[!t]
\centering
\includegraphics[width=0.28\textwidth]{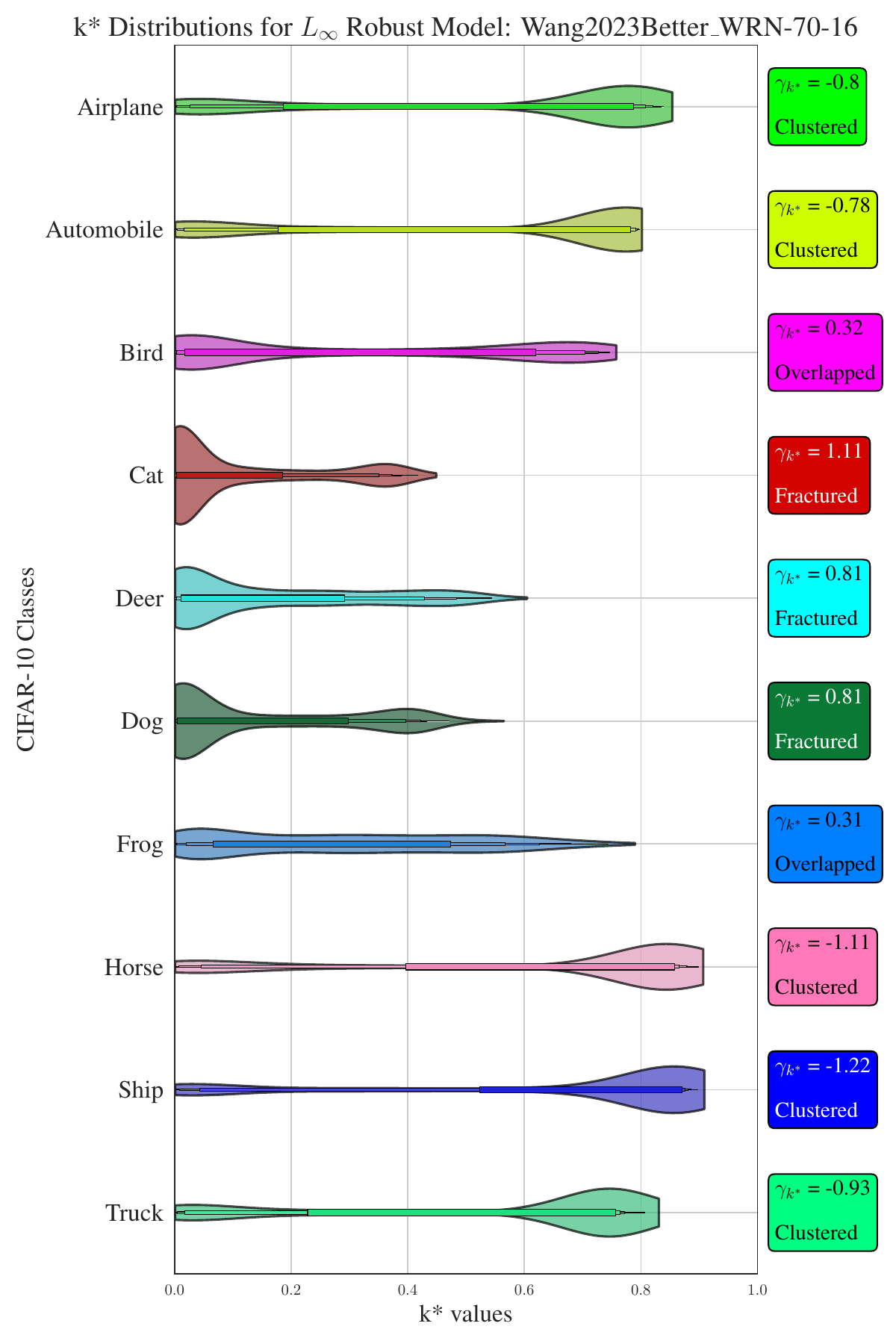}
\includegraphics[width=0.21\textwidth]{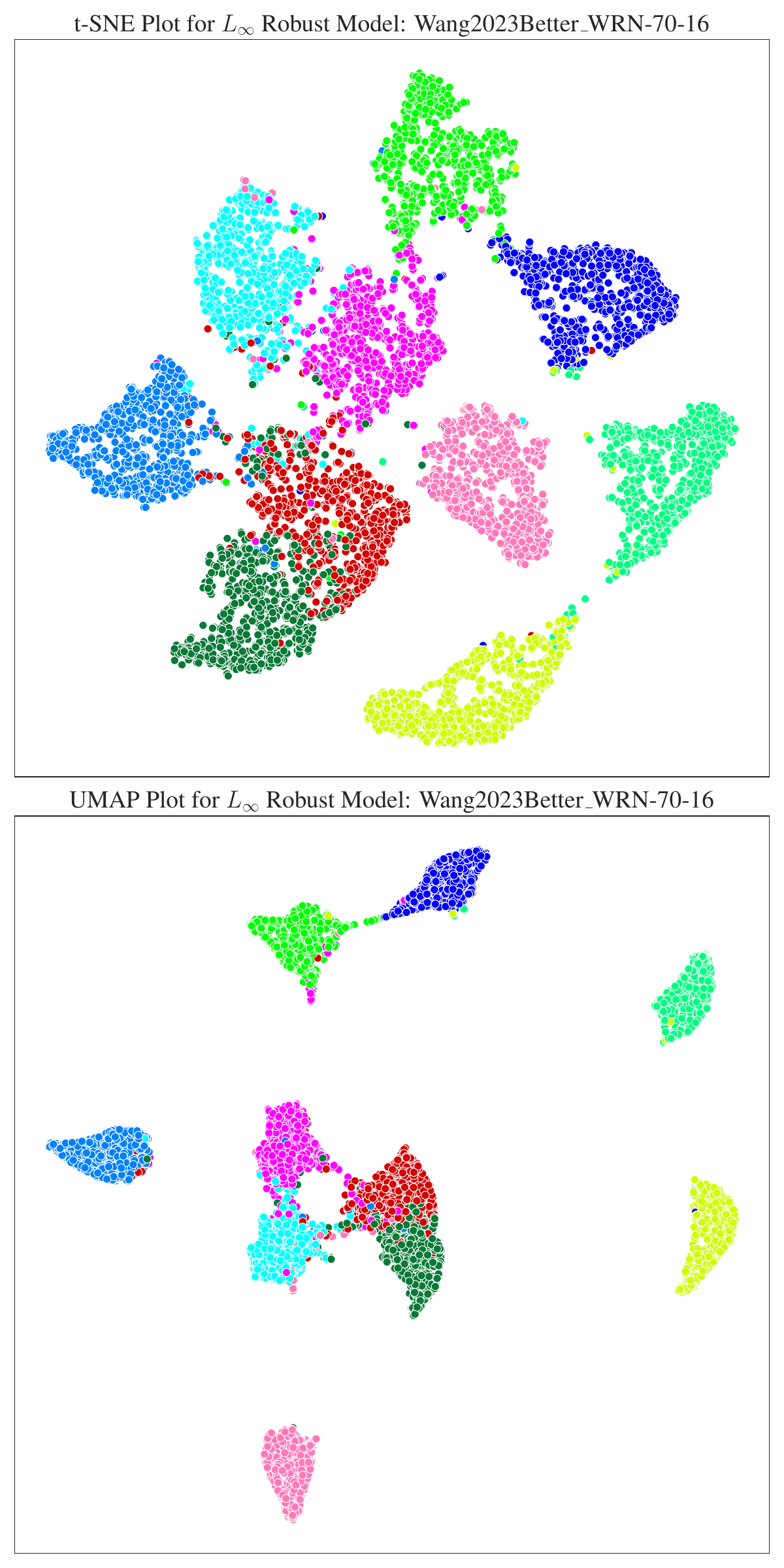}
\unskip\ \vrule\ 
\includegraphics[width=0.28\textwidth]{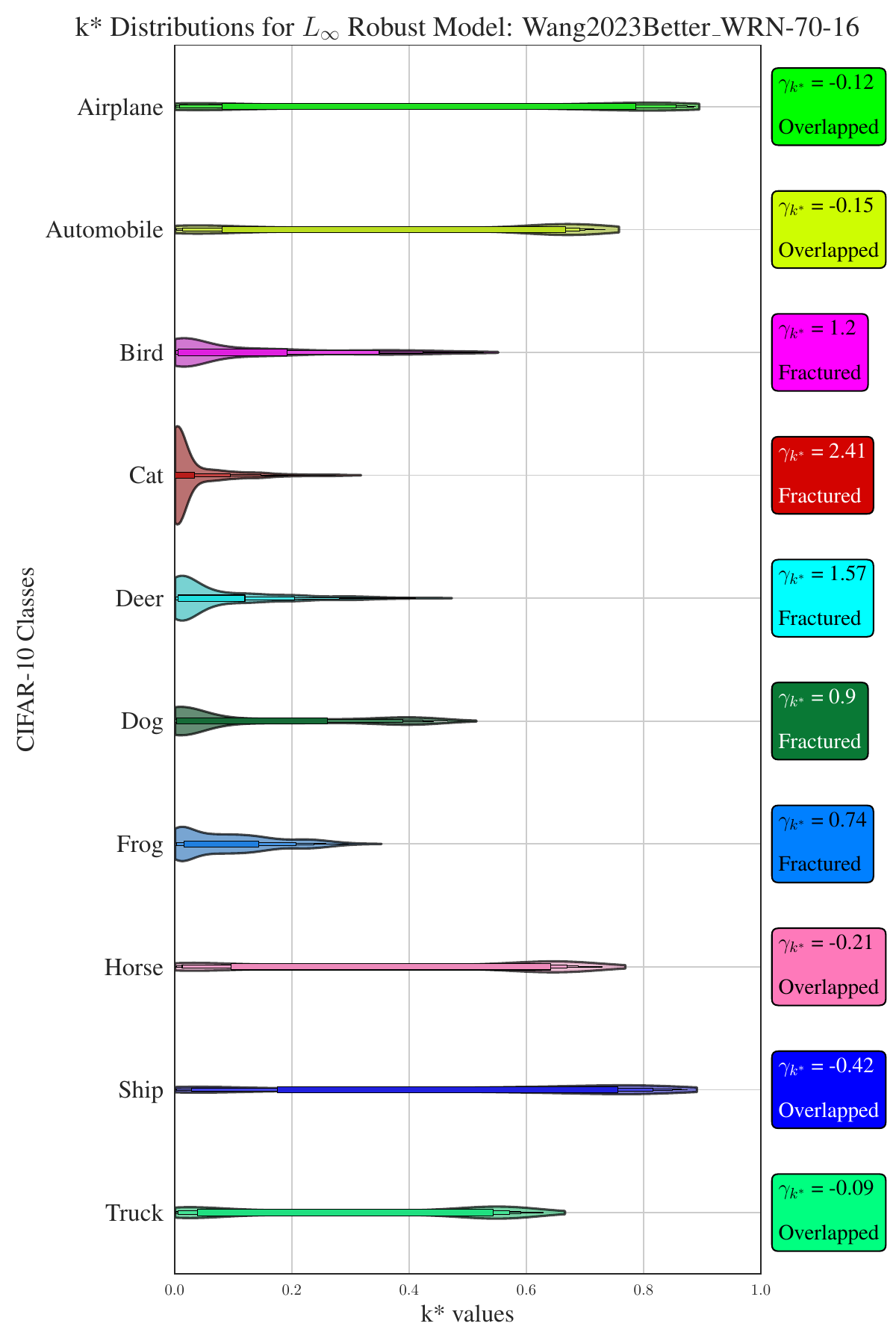}
\includegraphics[width=0.21\textwidth]{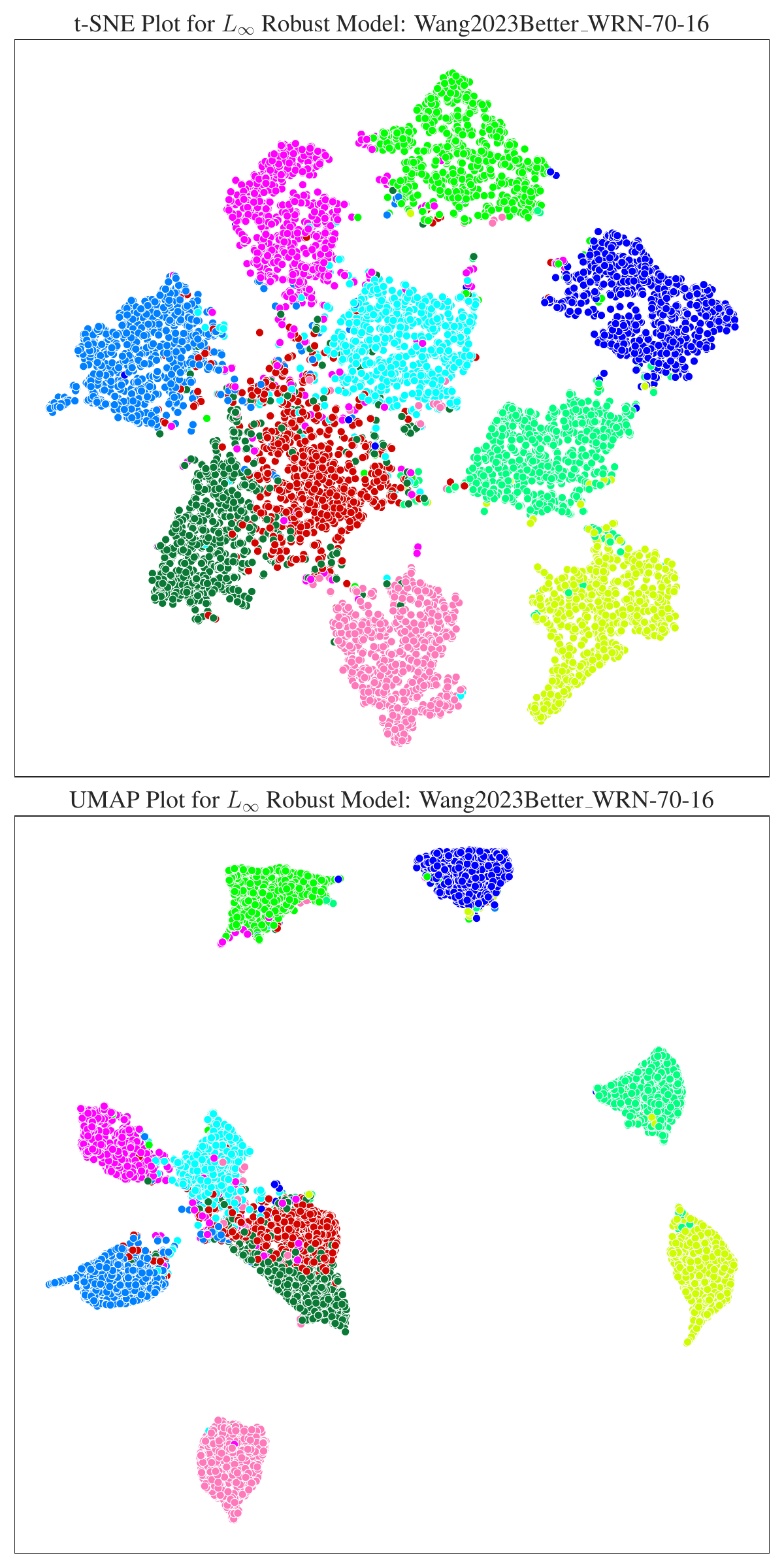}

\caption{
    Comparison of two variants of robust Wang2023Better\_WRN-70-16 models: a) $L_2$ Robust (\textit{Left}) and b) $L_\infty$ Robust (\textit{Right}) Models using k* Distribution, t-SNE and UMAP. 
    Note that information between the k* Distribution, UMAP, and t-SNE remains the same. 
    However, it is easier to compare the models by visually inspecting k* Distribution compared to t-SNE and UMAP. 
    This problem of comparing using t-SNE and UMAP becomes more prevalent as a number of concepts (classes) increases or the number of models increases. 
}
\label{fig:compare}
\end{figure*}

\begin{figure*}[!t]
\centering
\includegraphics[width=0.99\textwidth]{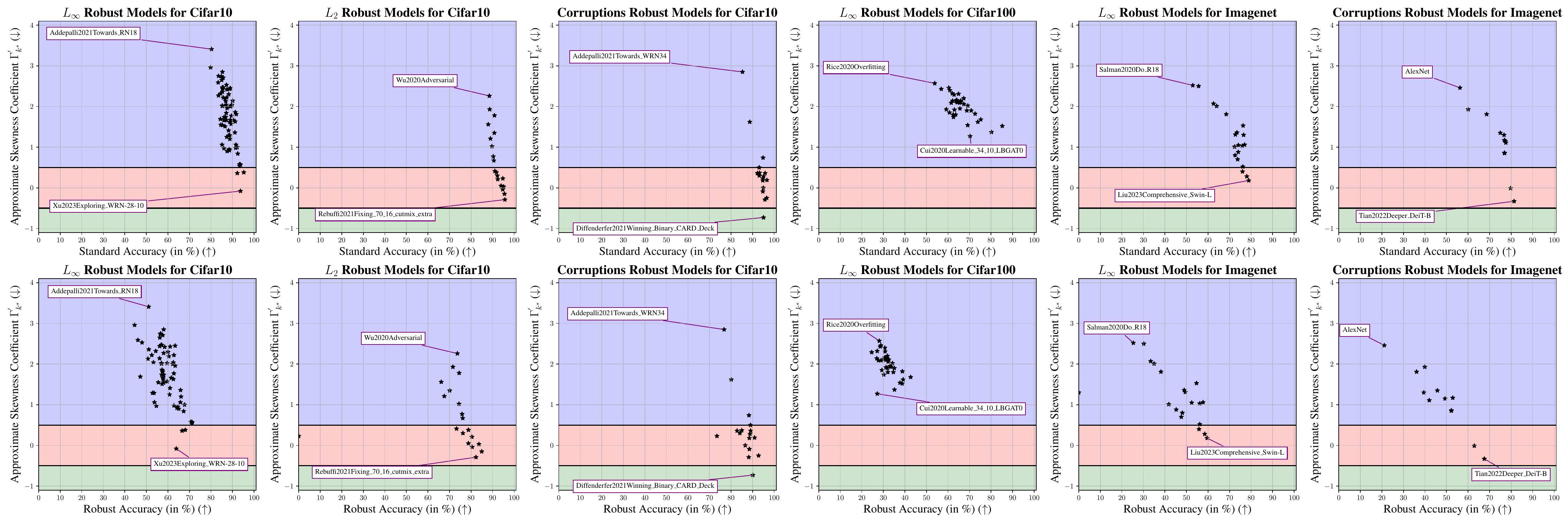}
\caption{
    Comparison of different robust models using the Approximate Skewness Coefficient $\Gamma^{'}_{k^*}$.\\
    \textit{Note that $\gamma_{k^*} > 0.5$ means the latent space is mostly fractured (Blue Region), $ -0.5 < \gamma_{k^*} < 0.5$ means the latent space is mostly overlapped (Red Region), while $\gamma_{k^*}< -0.5$ means the latent space is mostly clustered (Green Region).}
}
\label{fig:robust}
\end{figure*}

\section{Analyzing Robust Image Encoders}

\begin{table}[!t]
\centering
\resizebox{.99\columnwidth}{!}{
\begin{tabular}{l|rr|rrr}
    \toprule
    \textbf{Category of Models} & \textbf{$\Gamma_{k^*}$} & \textbf{$\Gamma^{'}_{k^*}$} &
    \textbf{\Fractured~Classes} & \textbf{\Overlapped~Classes} & \textbf{\Clustered~Classes} \\
    \midrule
    \multicolumn{6}{c}{\textbf{CIFAR-10}} \\
    \midrule
    Corruptions Robust  & 2.15 & 1.76 & 4.05 & 4.11 & 1.82 \\
    $L_2$ Robust        & 1.00 & 0.72 & 5.63 & 2.79 & 1.58 \\
    $L_\infty$ Robust   & 0.72 & 0.38 & 8.55 & 1.36 & 0.08 \\
    \midrule
    \multicolumn{6}{c}{\textbf{CIFAR-100}} \\
    \midrule
    $L_\infty$ Robust   & 3.70 & 1.99 & 94.28 & 5.17 & 0.55\\
    \midrule
    \multicolumn{4}{c}{\textbf{ImageNet-1k}} \\
    \midrule
    Corruptions Robust  & 2.31 & 1.14 & 756.83 & 160.00 & 83.16 \\
    $L_\infty$ Robust   & 2.59 & 1.21 & 760.10 & 158.25 & 81.65 \\
    \bottomrule
\end{tabular}
}
\caption{Comparison of the average number of classes \fractured~(\Fractured), \overlapped~(\Overlapped), and \clustered~(\Clustered) by different robust models.}
\label{tab:number}
\end{table}

\subsection{Experimental Setup}
In this study, we evaluate the robust vision models available at RobustBench Library \cite{croce2020robustbench} 
These models have been comprehensively evaluated through various adversarial attacks to be classified as robust. 
The repository offers pre-trained weights for these models, which we use for our analysis.

\subsection{Results}
\Figref{fig:compare} compares two variants of robust Wang2023Better\_WRN-70-16 model: an $L_2$ robust variant and an $L_\infty$ robust variant. 
Visual inspection reveals that understanding the latent space using t-SNE and UMAP is subjective, making it difficult to compare latent spaces using these methods. 
Furthermore, limited insights can be derived from this comparison.
In contrast, comparing the k* Distribution between the variants reveals a significant difference. 
Additionally, each class can be individually analyzed, providing further insights into individual classes. 
The k* Distribution allows us to compare and conclude that the $L_2$ robust variant of the model is better at evaluating individual concepts.
In contrast, the $L_\infty$ robust variant tends to fracture the latent space more.

To further determine the differences between the $L_2$ robust, $L_\infty$ robust, and Corruptions robust variants,  the k* distribution is computed across all robust models available in \cite{croce2020robustbench}. 
The average number of fractured, overlapped, and clustered classes is also computed. 
This analysis, reported in \Tableref{tab:number} shows that the number of fractured classes increases in the order of $L_\infty$ robust, $L_2$ robust, and Corruptions robust.

Additionally, when  $\Gamma_{k^*}$ and $\Gamma^{'}_{k^*}$ are computed for each model and averaged, a noticeable difference in quantifying the latent space is observed. 
We also notice a high variation in the individual skewness coefficient $\gamma_{i, k^*}$, accounting for the difference in approximation. 
However, $\Gamma^{'}_{k^*}$ is consistently less than $\Gamma_{k^*}$.
The individual analyses of the models are provided in the appendix.

To analyze the quality of the latent space with respect to performance, a comparison of the models based on the Average Skewness Coefficient $\Gamma^{'}_{k^*}$ over both Natural Accuracy and Robust Accuracy is visualized in \Figref{fig:robust}. 
The comparison includes $L_\infty$, $L_2$, and Corruptions robust models for CIFAR-10, $L_\infty$ robust models for CIFAR-100, and $L_\infty$ and Corruptions robust models for ImageNet-1k. The figure shows that as the model improves in both Natural Accuracy and Robust Accuracy, there is a degradation in the degree of fracturing evaluated using $\Gamma^{'}_{k^*}$.

This analysis demonstrates that as models become more robust, they better interpret the concepts (classes) by improving the clustering of the concept's samples. 
However, it is also noted that $\Gamma^{'}_{k^*} > 0$ for all $L_\infty$ robust and $L_2$ robust models, and most Corruptions robust models, indicating that while these methods improve clustering, the overall latent space remains fractured, preventing the formation of $n$ homogenous clusters, where $n$ is the number of concepts.

\section{Analyzing CLIP-based Image Encoders}

\subsection{Experimental Setup}
Open CLIP models, as provided by \cite{ilharco_gabriel_2021_5143773} are utilized to evaluate CLIP-based vision models in this study. 
These models have undergone extensive evaluation on various datasets, and the repository offers both evaluation scripts and pre-trained weights, which are leveraged for our analysis.
We evaluate the models on different datasets inspired by evaluation strategies defined by VTAB \cite{zhai2019large} and ELEVATER \cite{li2022elevater} encompassing multiple categories as described below,  

\begin{description}[style=nextline, leftmargin=*, itemsep=0.5em]

    \item[Natural Image Datasets]
    These datasets contain natural images captured using standard cameras for classical vision problems.
    The classes may represent Generic, Fine-Grained, or Abstract objects. 
    This group includes: 
    Caltech-101 \cite{caltech}, 
    CIFAR-10, \cite{cifar}, CIFAR-100 \cite{cifar},
    Country211 \cite{radford2021learning}
    Describable Textures (DTD) \cite{dtd}, 
    FGVC Aircraft \cite{fgvc}, 
    Food-101 \cite{food}, 
    GTSRB \cite{gtsrb}, 
    Oxford Flowers-102 \cite{flowers}, 
    Oxford IIIT-Pets \cite{pets}, 
    Pascal VOC2007 \cite{voc}, and
    Stanford Cars \cite{cars}. 

    \item[Specialized Image Datasets]
    These datasets contain images captured through specialist equipment for specialized problems. 
    It contains three subgroups of:
    a) Remote-sensing, consisting of datasets such as 
    Resisc45 \cite{resisc} and 
    Eurosat \cite{eurosat}, 
    b) Medical Images, consisting of datasets such as 
    Patch Camelyon \cite{patchcamelyon}.

    \item[Structured Image Datasets]
    These datasets assess comprehension of the structure of scene and contain datasets of a variety of tasks, such as 
    3D dataset like Clevr \cite{clevr}, 
    handwritten digits like MNIST \cite{mnist}, 
    optical character recognition like Rendered SST-2 \cite{radford2021learning}, and 
    frames captured from a car like KITTI Dataset \cite{kitti}.

   \item[ImageNet-1k-like Image Datasets]
   These datasets belong to the same domain as the original 
   ImageNet-1k \cite{imagenet} like, 
   ImageNet-Sketch \cite{imagenetsketch}, 
   ImageNet-v2 \cite{imagenetv2}, 
   ImageNet-A \cite{imagenetao}, ImageNet-O \cite{imagenetao}, 
   ImageNet-R \cite{imagenetr}, 
   ObjectNet \cite{objectnet}.

\end{description}

\begin{figure*}[!t]
\centering
\includegraphics[width=0.99\textwidth]{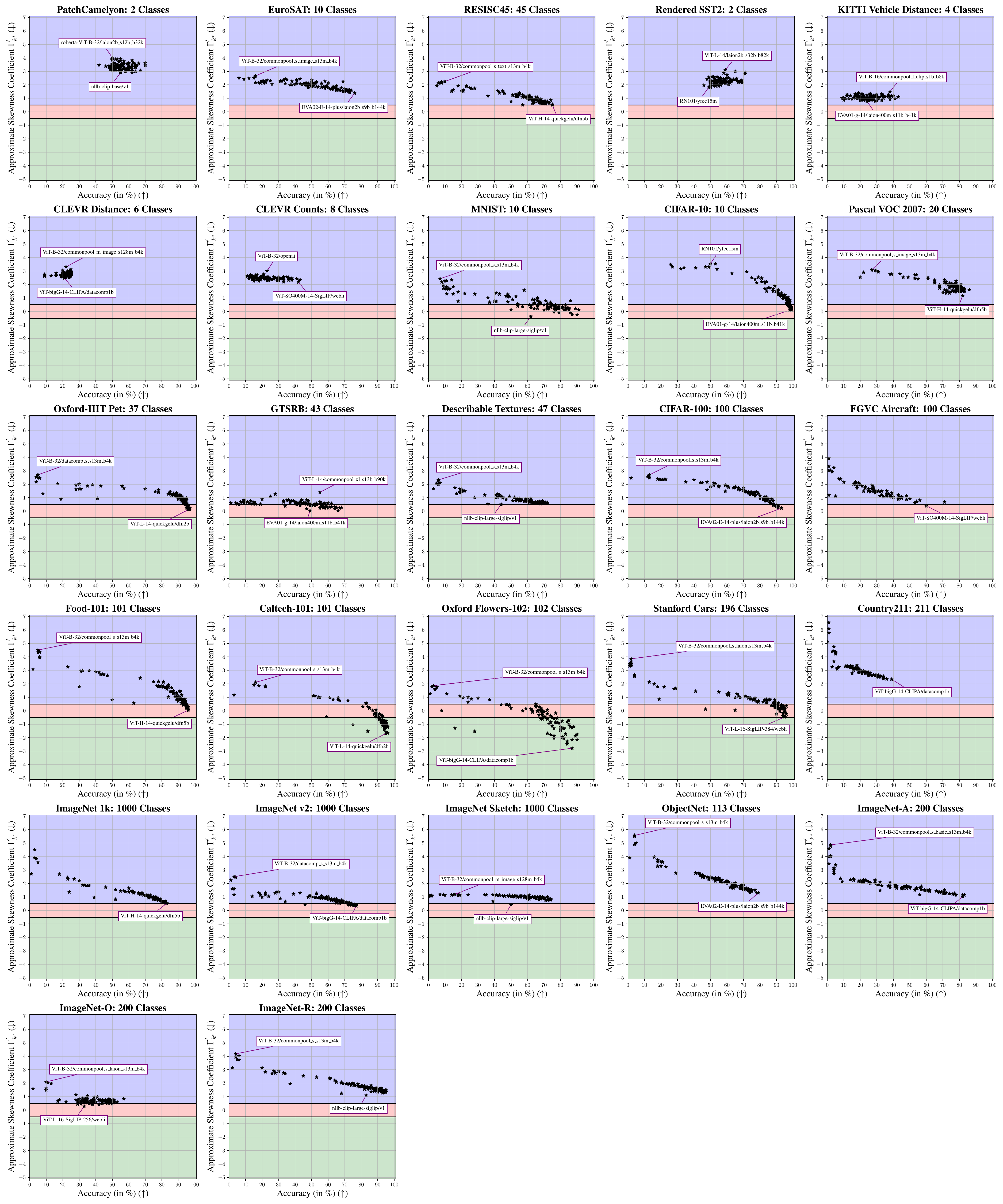}
\caption{
    Comparison of different Open-CLIP models using the Approximate Skewness Coefficient $\Gamma^{'}_{k^*}$ for multiple datasets.\\
    \textit{Note that $\gamma_{k^*} > 0.5$ means the latent space is mostly fractured (Blue Region), $ -0.5 < \gamma_{k^*} < 0.5$ means the latent space is mostly overlapped (Red Region), while $\gamma_{k^*}< -0.5$ means the latent space is mostly clustered (Green Region).}
}
\label{fig:clip}
\end{figure*}

\subsection{Results}
\Figref{fig:clip} a comparison of different Open CLIP models is presented using the Approximate Skewness Coefficient $\Gamma^{'}_{k^*}$. 
Comparision using True Skewness Coefficient $\Gamma_{k^*}$ is presented in the appendix.
The x-axis displays the accuracy of the models, allowing us to correlate performance with the quality of the latent space. 
A general trend emerges where a lower Approximate Skewness Coefficient  $\Gamma^{'}_{k^*}$ is associated with better accuracy in zero-shot classification across the evaluated datasets.

However, this trend does not apply universally to all datasets. 
For structured datasets, the evaluated models consistently fracture the latent space similarly, indicating that current training strategies do not enhance concept understanding for these datasets. 
Results for Rendered SST 2 further reveal that vision models struggle with clustering the characters and semantic meaning, indicating why vision models struggle to generate the text when employed in image generation.

Datasets like Patch Camelyon, GTSRB, ImageNet-Sketch, and ImageNet-O also deviate from this pattern. 
These datasets have minimal deviation in Approximate Skewness Coefficient $\Gamma^{'}_{k^*}$, indicating that current models do not improve the clustering of concepts for these datasets. 
We also note that while datasets like Patch Camelyon are highly fractured, datasets like GTSRB, ImageNet-Sketch, and ImageNet-O have less degree of fracturing, but the latent space is highly overlapped for these samples.  
Improvement in performance for these datasets, therefore, can be attributed to better projection to linearly separate the concepts rather than the vision model's intrinsic property. 

\section{Conclusion}
In this study, the k* Distribution is utilized to directly examine the latent space of various types of vision models (such as robust and CLIP-based), revealing insights into how individual concepts are structured in these models' latent spaces. 
By introducing skewness-based metrics, both true and approximate, the study quantifies the quality of these latent spaces.
The findings highlight that current vision models often fragment the distributions of individual concepts within the latent space. 
However, as models improve in generalization and robustness, the degree of fracturing decreases. 
This suggests that better generalization and robustness are associated with a more coherent clustering of concepts in the latent space. 
The quantification of the analysis of k* Distribution offers a direct and interpretable approach for comparing latent spaces, establishing a clear relationship between a model's generalization, robustness, and the quality of its latent space.

{
\bibliography{aaai25}

\begin{thebibliography}{48}
\providecommand{\natexlab}[1]{#1}

\bibitem[{Amid and Warmuth(2022)}]{amid2022TriMap}
Amid, E.; and Warmuth, M.~K. 2022.
\newblock {{TriMap}}: {{Large-scale Dimensionality Reduction Using Triplets}}.
\newblock arxiv:1910.00204.

\bibitem[{Arendt et~al.(2020)Arendt, Nur, Huang, Fair, and
  Dou}]{arendt2020Parallel}
Arendt, D.~L.; Nur, N.; Huang, Z.; Fair, G.; and Dou, W. 2020.
\newblock Parallel Embeddings: A Visualization Technique for Contrasting
  Learned Representations.
\newblock In \emph{Proceedings of the 25th {{International Conference}} on
  {{Intelligent User Interfaces}}}, {{IUI}} '20, 259--274. {New York, NY, USA}:
  {Association for Computing Machinery}.
\newblock ISBN 978-1-4503-7118-6.

\bibitem[{Barbu et~al.(2019)Barbu, Mayo, Alverio, Luo, Wang, Gutfreund,
  Tenenbaum, and Katz}]{objectnet}
Barbu, A.; Mayo, D.; Alverio, J.; Luo, W.; Wang, C.; Gutfreund, D.; Tenenbaum,
  J.; and Katz, B. 2019.
\newblock Objectnet: A large-scale bias-controlled dataset for pushing the
  limits of object recognition models.
\newblock \emph{Advances in neural information processing systems}, 32.

\bibitem[{Boggust, Carter, and Satyanarayan(2022)}]{boggust2022Embedding}
Boggust, A.; Carter, B.; and Satyanarayan, A. 2022.
\newblock Embedding {{Comparator}}: {{Visualizing Differences}} in {{Global
  Structure}} and {{Local Neighborhoods}} via {{Small Multiples}}.
\newblock In \emph{27th {{International Conference}} on {{Intelligent User
  Interfaces}}}, {{IUI}} '22, 746--766. {New York, NY, USA}: {Association for
  Computing Machinery}.
\newblock ISBN 978-1-4503-9144-3.

\bibitem[{Bossard, Guillaumin, and Van~Gool(2014)}]{food}
Bossard, L.; Guillaumin, M.; and Van~Gool, L. 2014.
\newblock Food-101--mining discriminative components with random forests.
\newblock In \emph{Computer vision--ECCV 2014: 13th European conference,
  zurich, Switzerland, September 6-12, 2014, proceedings, part VI 13},
  446--461. Springer.

\bibitem[{Carter et~al.(2019)Carter, Armstrong, Schubert, Johnson, and
  Olah}]{activation-atlas}
Carter, S.; Armstrong, Z.; Schubert, L.; Johnson, I.; and Olah, C. 2019.
\newblock Activation Atlas.
\newblock \emph{Distill}.
\newblock Https://distill.pub/2019/activation-atlas.

\bibitem[{Cheng, Han, and Lu(2017)}]{resisc}
Cheng, G.; Han, J.; and Lu, X. 2017.
\newblock Remote sensing image scene classification: Benchmark and state of the
  art.
\newblock \emph{Proceedings of the IEEE}, 105(10): 1865--1883.

\bibitem[{Cherti et~al.(2023)Cherti, Beaumont, Wightman, Wortsman, Ilharco,
  Gordon, Schuhmann, Schmidt, and Jitsev}]{cherti2023reproducible}
Cherti, M.; Beaumont, R.; Wightman, R.; Wortsman, M.; Ilharco, G.; Gordon, C.;
  Schuhmann, C.; Schmidt, L.; and Jitsev, J. 2023.
\newblock Reproducible scaling laws for contrastive language-image learning.
\newblock In \emph{Proceedings of the IEEE/CVF Conference on Computer Vision
  and Pattern Recognition}, 2818--2829.

\bibitem[{Cimpoi et~al.(2014)Cimpoi, Maji, Kokkinos, Mohamed, and
  Vedaldi}]{dtd}
Cimpoi, M.; Maji, S.; Kokkinos, I.; Mohamed, S.; and Vedaldi, A. 2014.
\newblock Describing textures in the wild.
\newblock In \emph{Proceedings of the IEEE conference on computer vision and
  pattern recognition}, 3606--3613.

\bibitem[{Coifman and Lafon(2006)}]{coifman2006diffusion}
Coifman, R.~R.; and Lafon, S. 2006.
\newblock Diffusion maps.
\newblock \emph{Applied and computational harmonic analysis}, 21(1): 5--30.

\bibitem[{Croce et~al.(2020)Croce, Andriushchenko, Sehwag, Debenedetti,
  Flammarion, Chiang, Mittal, and Hein}]{croce2020robustbench}
Croce, F.; Andriushchenko, M.; Sehwag, V.; Debenedetti, E.; Flammarion, N.;
  Chiang, M.; Mittal, P.; and Hein, M. 2020.
\newblock RobustBench: a standardized adversarial robustness benchmark.
\newblock \emph{arXiv preprint arXiv:2010.09670}.

\bibitem[{Cutura et~al.(2020)Cutura, Aupetit, Fekete, and
  Sedlmair}]{cutura2020Comparing}
Cutura, R.; Aupetit, M.; Fekete, J.-D.; and Sedlmair, M. 2020.
\newblock Comparing and {{Exploring High-Dimensional Data}} with
  {{Dimensionality Reduction Algorithms}} and {{Matrix Visualizations}}.
\newblock In \emph{Proceedings of the {{International Conference}} on
  {{Advanced Visual Interfaces}}}, {{AVI}} '20, 1--9. {New York, NY, USA}:
  {Association for Computing Machinery}.
\newblock ISBN 978-1-4503-7535-1.

\bibitem[{Deng et~al.(2009)Deng, Dong, Socher, Li, Li, and Fei-Fei}]{imagenet}
Deng, J.; Dong, W.; Socher, R.; Li, L.-J.; Li, K.; and Fei-Fei, L. 2009.
\newblock Imagenet: A large-scale hierarchical image database.
\newblock In \emph{2009 IEEE conference on computer vision and pattern
  recognition}, 248--255. Ieee.

\bibitem[{Dosovitskiy et~al.(2020)Dosovitskiy, Beyer, Kolesnikov, Weissenborn,
  Zhai, Unterthiner, Dehghani, Minderer, Heigold, Gelly
  et~al.}]{dosovitskiy2020image}
Dosovitskiy, A.; Beyer, L.; Kolesnikov, A.; Weissenborn, D.; Zhai, X.;
  Unterthiner, T.; Dehghani, M.; Minderer, M.; Heigold, G.; Gelly, S.; et~al.
  2020.
\newblock An image is worth 16x16 words: Transformers for image recognition at
  scale.
\newblock \emph{arXiv preprint arXiv:2010.11929}.

\bibitem[{Everingham et~al.(2010)Everingham, Van~Gool, Williams, Winn, and
  Zisserman}]{voc}
Everingham, M.; Van~Gool, L.; Williams, C.~K.; Winn, J.; and Zisserman, A.
  2010.
\newblock The pascal visual object classes (voc) challenge.
\newblock \emph{International journal of computer vision}, 88: 303--338.

\bibitem[{Fei-Fei, Fergus, and Perona(2004)}]{caltech}
Fei-Fei, L.; Fergus, R.; and Perona, P. 2004.
\newblock Learning generative visual models from few training examples: An
  incremental bayesian approach tested on 101 object categories.
\newblock In \emph{2004 conference on computer vision and pattern recognition
  workshop}, 178--178. IEEE.

\bibitem[{Geiger et~al.(2013)Geiger, Lenz, Stiller, and Urtasun}]{kitti}
Geiger, A.; Lenz, P.; Stiller, C.; and Urtasun, R. 2013.
\newblock Vision meets robotics: The kitti dataset.
\newblock \emph{The International Journal of Robotics Research}, 32(11):
  1231--1237.

\bibitem[{Gleicher et~al.(2011)Gleicher, Albers, Walker, Jusufi, Hansen, and
  Roberts}]{gleicher2011Visual}
Gleicher, M.; Albers, D.; Walker, R.; Jusufi, I.; Hansen, C.~D.; and Roberts,
  J.~C. 2011.
\newblock Visual Comparison for Information Visualization.
\newblock \emph{Information Visualization}, 10(4): 289--309.

\bibitem[{Helber et~al.(2019)Helber, Bischke, Dengel, and Borth}]{eurosat}
Helber, P.; Bischke, B.; Dengel, A.; and Borth, D. 2019.
\newblock Eurosat: A novel dataset and deep learning benchmark for land use and
  land cover classification.
\newblock \emph{IEEE Journal of Selected Topics in Applied Earth Observations
  and Remote Sensing}, 12(7): 2217--2226.

\bibitem[{Hendrycks et~al.(2021{\natexlab{a}})Hendrycks, Basart, Mu, Kadavath,
  Wang, Dorundo, Desai, Zhu, Parajuli, Guo, Song, Steinhardt, and
  Gilmer}]{imagenetr}
Hendrycks, D.; Basart, S.; Mu, N.; Kadavath, S.; Wang, F.; Dorundo, E.; Desai,
  R.; Zhu, T.; Parajuli, S.; Guo, M.; Song, D.; Steinhardt, J.; and Gilmer, J.
  2021{\natexlab{a}}.
\newblock The Many Faces of Robustness: A Critical Analysis of
  Out-of-Distribution Generalization.
\newblock \emph{ICCV}.

\bibitem[{Hendrycks et~al.(2021{\natexlab{b}})Hendrycks, Zhao, Basart,
  Steinhardt, and Song}]{imagenetao}
Hendrycks, D.; Zhao, K.; Basart, S.; Steinhardt, J.; and Song, D.
  2021{\natexlab{b}}.
\newblock Natural Adversarial Examples.
\newblock \emph{CVPR}.

\bibitem[{Hotelling(1933)}]{hotelling1933analysis}
Hotelling, H. 1933.
\newblock Analysis of a complex of statistical variables into principal
  components.
\newblock \emph{Journal of educational psychology}, 24(6): 417.

\bibitem[{Ilharco et~al.(2021)Ilharco, Wortsman, Wightman, Gordon, Carlini,
  Taori, Dave, Shankar, Namkoong, Miller, Hajishirzi, Farhadi, and
  Schmidt}]{ilharco_gabriel_2021_5143773}
Ilharco, G.; Wortsman, M.; Wightman, R.; Gordon, C.; Carlini, N.; Taori, R.;
  Dave, A.; Shankar, V.; Namkoong, H.; Miller, J.; Hajishirzi, H.; Farhadi, A.;
  and Schmidt, L. 2021.
\newblock OpenCLIP.
\newblock If you use this software, please cite it as below.

\bibitem[{Jia et~al.(2021)Jia, Yang, Xia, Chen, Parekh, Pham, Le, Sung, Li, and
  Duerig}]{jia2021scaling}
Jia, C.; Yang, Y.; Xia, Y.; Chen, Y.-T.; Parekh, Z.; Pham, H.; Le, Q.; Sung,
  Y.-H.; Li, Z.; and Duerig, T. 2021.
\newblock Scaling up visual and vision-language representation learning with
  noisy text supervision.
\newblock In \emph{International conference on machine learning}, 4904--4916.
  PMLR.

\bibitem[{Johnson et~al.(2017)Johnson, Hariharan, Van Der~Maaten, Fei-Fei,
  Lawrence~Zitnick, and Girshick}]{clevr}
Johnson, J.; Hariharan, B.; Van Der~Maaten, L.; Fei-Fei, L.; Lawrence~Zitnick,
  C.; and Girshick, R. 2017.
\newblock Clevr: A diagnostic dataset for compositional language and elementary
  visual reasoning.
\newblock In \emph{Proceedings of the IEEE conference on computer vision and
  pattern recognition}, 2901--2910.

\bibitem[{Kornblith, Shlens, and Le(2019)}]{kornblith2019better}
Kornblith, S.; Shlens, J.; and Le, Q.~V. 2019.
\newblock Do better imagenet models transfer better?
\newblock In \emph{Proceedings of the IEEE/CVF conference on computer vision
  and pattern recognition}, 2661--2671.

\bibitem[{Kotyan, Ueda, and Vargas(2023)}]{kotyan2023k}
Kotyan, S.; Ueda, T.; and Vargas, D.~V. 2023.
\newblock k* Distribution: Evaluating the Latent Space of Deep Neural Networks
  using Local Neighborhood Analysis.
\newblock \emph{arXiv preprint arXiv:2312.04024}.

\bibitem[{Krause et~al.(2013)Krause, Stark, Deng, and Fei-Fei}]{cars}
Krause, J.; Stark, M.; Deng, J.; and Fei-Fei, L. 2013.
\newblock 3d object representations for fine-grained categorization.
\newblock In \emph{Proceedings of the IEEE international conference on computer
  vision workshops}, 554--561.

\bibitem[{Krizhevsky, Hinton et~al.(2009)}]{cifar}
Krizhevsky, A.; Hinton, G.; et~al. 2009.
\newblock Learning multiple layers of features from tiny images.

\bibitem[{Kruskal(1964)}]{kruskal1964multidimensional}
Kruskal, J.~B. 1964.
\newblock Multidimensional scaling by optimizing goodness of fit to a nonmetric
  hypothesis.
\newblock \emph{Psychometrika}, 29(1): 1--27.

\bibitem[{LeCun, Cortes, and Burges(2010)}]{mnist}
LeCun, Y.; Cortes, C.; and Burges, C. 2010.
\newblock MNIST handwritten digit database.
\newblock \emph{ATT Labs [Online]. Available:
  http://yann.lecun.com/exdb/mnist}, 2.

\bibitem[{Li et~al.(2022)Li, Liu, Li, Zhang, Aneja, Yang, Jin, Hu, Liu, Lee
  et~al.}]{li2022elevater}
Li, C.; Liu, H.; Li, L.; Zhang, P.; Aneja, J.; Yang, J.; Jin, P.; Hu, H.; Liu,
  Z.; Lee, Y.~J.; et~al. 2022.
\newblock Elevater: A benchmark and toolkit for evaluating language-augmented
  visual models.
\newblock \emph{Advances in Neural Information Processing Systems}, 35:
  9287--9301.

\bibitem[{Maaten and Hinton(2008)}]{maaten2008visualizing}
Maaten, L. v.~d.; and Hinton, G. 2008.
\newblock Visualizing data using t-SNE.
\newblock \emph{Journal of Machine Learning Research}, 9(Nov): 2579--2605.

\bibitem[{Mahendran and Vedaldi(2015)}]{visualizing_activation_1}
Mahendran, A.; and Vedaldi, A. 2015.
\newblock Understanding deep image representations by inverting them.
\newblock In \emph{Proceedings of the IEEE conference on computer vision and
  pattern recognition}, 5188--5196.

\bibitem[{Maji et~al.(2013)Maji, Rahtu, Kannala, Blaschko, and Vedaldi}]{fgvc}
Maji, S.; Rahtu, E.; Kannala, J.; Blaschko, M.; and Vedaldi, A. 2013.
\newblock Fine-grained visual classification of aircraft.
\newblock \emph{arXiv preprint arXiv:1306.5151}.

\bibitem[{{McInnes}, {Healy}, and {Melville}(2018)}]{2018arXivUMAP}
{McInnes}, L.; {Healy}, J.; and {Melville}, J. 2018.
\newblock {UMAP: Uniform Manifold Approximation and Projection for Dimension
  Reduction}.
\newblock \emph{ArXiv e-prints}.

\bibitem[{Nilsback and Zisserman(2008)}]{flowers}
Nilsback, M.-E.; and Zisserman, A. 2008.
\newblock Automated flower classification over a large number of classes.
\newblock In \emph{2008 Sixth Indian conference on computer vision, graphics \&
  image processing}, 722--729. IEEE.

\bibitem[{Olah et~al.(2018)Olah, Satyanarayan, Johnson, Carter, Schubert, Ye,
  and Mordvintsev}]{visualizing_activation_2}
Olah, C.; Satyanarayan, A.; Johnson, I.; Carter, S.; Schubert, L.; Ye, K.; and
  Mordvintsev, A. 2018.
\newblock The building blocks of interpretability.
\newblock \emph{Distill}, 3(3): e10.

\bibitem[{Parkhi et~al.(2012)Parkhi, Vedaldi, Zisserman, and Jawahar}]{pets}
Parkhi, O.~M.; Vedaldi, A.; Zisserman, A.; and Jawahar, C. 2012.
\newblock Cats and dogs.
\newblock In \emph{2012 IEEE conference on computer vision and pattern
  recognition}, 3498--3505. IEEE.

\bibitem[{Radford et~al.(2021)Radford, Kim, Hallacy, Ramesh, Goh, Agarwal,
  Sastry, Askell, Mishkin, Clark et~al.}]{radford2021learning}
Radford, A.; Kim, J.~W.; Hallacy, C.; Ramesh, A.; Goh, G.; Agarwal, S.; Sastry,
  G.; Askell, A.; Mishkin, P.; Clark, J.; et~al. 2021.
\newblock Learning transferable visual models from natural language
  supervision.
\newblock In \emph{International conference on machine learning}, 8748--8763.
  PMLR.

\bibitem[{Recht et~al.(2019)Recht, Roelofs, Schmidt, and Shankar}]{imagenetv2}
Recht, B.; Roelofs, R.; Schmidt, L.; and Shankar, V. 2019.
\newblock Do imagenet classifiers generalize to imagenet?
\newblock In \emph{International conference on machine learning}, 5389--5400.
  PMLR.

\bibitem[{Sivaraman, Wu, and Perer(2022)}]{sivaraman2022Emblaze}
Sivaraman, V.; Wu, Y.; and Perer, A. 2022.
\newblock Emblaze: {{Illuminating Machine Learning Representations}} through
  {{Interactive Comparison}} of {{Embedding Spaces}}.
\newblock In \emph{27th {{International Conference}} on {{Intelligent User
  Interfaces}}}, {{IUI}} '22, 418--432. {New York, NY, USA}: {Association for
  Computing Machinery}.
\newblock ISBN 978-1-4503-9144-3.

\bibitem[{Stallkamp et~al.(2012)Stallkamp, Schlipsing, Salmen, and
  Igel}]{gtsrb}
Stallkamp, J.; Schlipsing, M.; Salmen, J.; and Igel, C. 2012.
\newblock Man vs. computer: Benchmarking machine learning algorithms for
  traffic sign recognition.
\newblock \emph{Neural networks}, 32: 323--332.

\bibitem[{Veeling et~al.(2018)Veeling, Linmans, Winkens, Cohen, and
  Welling}]{patchcamelyon}
Veeling, B.~S.; Linmans, J.; Winkens, J.; Cohen, T.; and Welling, M. 2018.
\newblock Rotation equivariant CNNs for digital pathology.
\newblock In \emph{Medical Image Computing and Computer Assisted
  Intervention--MICCAI 2018: 21st International Conference, Granada, Spain,
  September 16-20, 2018, Proceedings, Part II 11}, 210--218. Springer.

\bibitem[{Wang et~al.(2019)Wang, Ge, Lipton, and Xing}]{imagenetsketch}
Wang, H.; Ge, S.; Lipton, Z.; and Xing, E.~P. 2019.
\newblock Learning Robust Global Representations by Penalizing Local Predictive
  Power.
\newblock In \emph{Advances in Neural Information Processing Systems},
  10506--10518.

\bibitem[{Wortsman et~al.(2022)Wortsman, Ilharco, Kim, Li, Kornblith, Roelofs,
  Lopes, Hajishirzi, Farhadi, Namkoong et~al.}]{wortsman2022robust}
Wortsman, M.; Ilharco, G.; Kim, J.~W.; Li, M.; Kornblith, S.; Roelofs, R.;
  Lopes, R.~G.; Hajishirzi, H.; Farhadi, A.; Namkoong, H.; et~al. 2022.
\newblock Robust fine-tuning of zero-shot models.
\newblock In \emph{Proceedings of the IEEE/CVF conference on computer vision
  and pattern recognition}, 7959--7971.

\bibitem[{Zhai et~al.(2019)Zhai, Puigcerver, Kolesnikov, Ruyssen, Riquelme,
  Lucic, Djolonga, Pinto, Neumann, Dosovitskiy et~al.}]{zhai2019large}
Zhai, X.; Puigcerver, J.; Kolesnikov, A.; Ruyssen, P.; Riquelme, C.; Lucic, M.;
  Djolonga, J.; Pinto, A.~S.; Neumann, M.; Dosovitskiy, A.; et~al. 2019.
\newblock A large-scale study of representation learning with the visual task
  adaptation benchmark.
\newblock \emph{arXiv preprint arXiv:1910.04867}.

\bibitem[{Zhou et~al.(2022)Zhou, Yang, Loy, and Liu}]{zhou2022learning}
Zhou, K.; Yang, J.; Loy, C.~C.; and Liu, Z. 2022.
\newblock Learning to prompt for vision-language models.
\newblock \emph{International Journal of Computer Vision}, 130(9): 2337--2348.

\end{thebibliography}
}
|
\end{document}